\documentclass{INTERSPEECH2023}
\usepackage[table,xcdraw]{xcolor}
\usepackage{xcolor}
\usepackage{adjustbox}
\usepackage{hyperref}
\usepackage{url}

\interspeechcameraready

\title{How to Estimate Model Transferability of Pre-Trained Speech Models?}
\name{Zih-Ching Chen$^1$, Chao-Han Huck Yang$^{2^{*},3}$\thanks{$^{*}$Work was mainly done during a research intern at Google.}, Bo Li$^2$, Yu Zhang$^2$ \\ Nanxin Chen$^2$, Shuo-Yiin Chang$^2$, Rohit Prabhavalkar$^2$, Hung-yi Lee$^1$, Tara N. Sainath$^2$}
\address{
  $^1$National Taiwan University,  Taiwan~~~ 
  $^2$Google, USA~~~
  $^3$Georgia Tech, USA}
\email{$\{$r09942176, hungyilee$\}$@ntu.edu.tw;~huckiyang@gatech.edu; $\{$boboli,ngyuzh$\}$@google.com}

\begin{document}
\maketitle
\begin{abstract}
In this work, we introduce a ``score-based assessment'' framework for estimating the transferability of pre-trained speech models (PSMs) for fine-tuning target tasks. We leverage upon two representation theories, Bayesian likelihood estimation and optimal transport, to generate rank scores for the PSM candidates using the extracted representations. Our framework efficiently computes transferability scores without actual fine-tuning of candidate models or layers by making a temporal independent hypothesis. We evaluate some popular supervised speech models (e.g., Conformer RNN-Transducer) and self-supervised speech models (e.g., HuBERT) in cross-layer and cross-model settings using public data.
Experimental results show a high Spearman's rank correlation and low $p$-value between our estimation framework and fine-tuning ground truth.
Our proposed transferability framework requires less computational time and resources, making it a resource-saving and time-efficient approach for tuning speech foundation models.

\end{abstract}

\section{Introduction}
In recent years, large-scale pre-trained neural networks, also known as Foundation Models~\cite{li2023efficient, zhang2022bigssl, radford2022robust, bommasani2021opportunities}, have demonstrated numerous benefits in various fields. One of these benefits includes the application of supervised learning \cite{raffel2020exploring} or self-supervised learning (SSL) models \cite{devlin2018bert, baevski2020wav2vec, hsu2021hubert} trained on few-shot adaptation~\cite{xiao2021scaling}~or continuous learning~\cite{kessler2021continual}~from pre-training. FMs have shown strong generalization and zero-shot learning abilities~\cite{brown2020language, ye2021towards} by learning representations, making them a popular choice for speech processing tasks~\cite{zhang2022bigssl, radford2022robust, yu2016automatic}. These representations can then be utilized by downstream models for specific task-related feature extraction~\cite{yang2021superb}. However, fine-tuning separate FMs for multiple downstream tasks can be computationally expensive and resource-intensive due to the size of the FMs. Even in partial model tuning settings, determining where and how to insert parameter-efficient learning modules such as residual adapters~\cite{chen2023exploring, chen2022chapter} and neural reprogramming~\cite{yang2021voice2series} relies heavily on hand-crafted expertise.

\begin{figure}[ht!]
    \centering
    \includegraphics[width=0.42\textwidth]{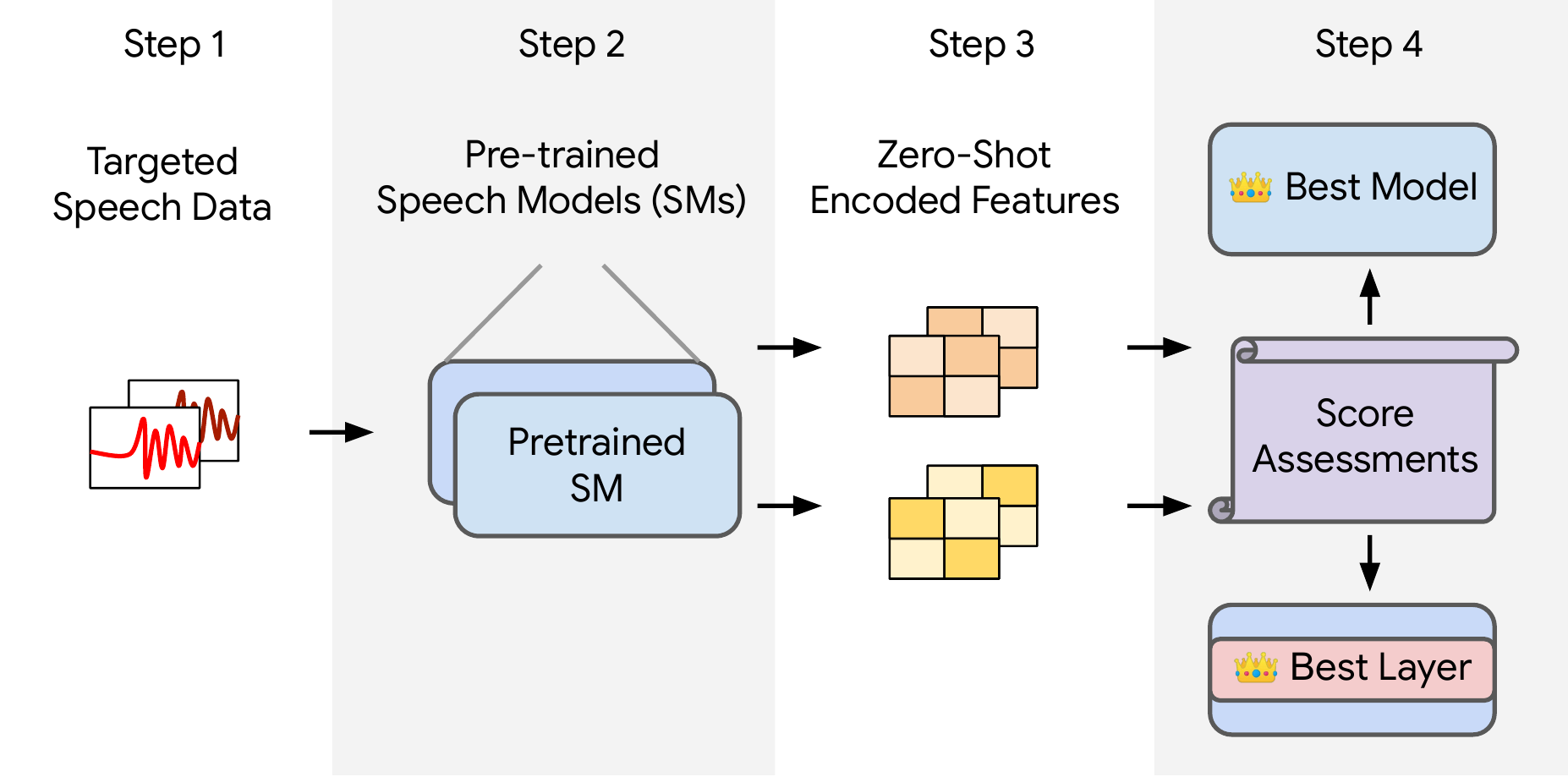}
    \caption{A step-by-step illustration of the proposed framework for providing scores for assessing the best pre-trained model and the best layer for transfer learning with speech data.}
    \label{fig:overall_diagram}
\end{figure}

In this work, we aim to address the aforementioned challenges of determining the best FM or candidate layer for tuning by \textit{introducing a model transfer assessment framework for speech processing} applications. As illustrated in Figure~\ref{fig:overall_diagram}, our process consists of four steps: step 1 and step 2 involve collecting target speech data and pre-trained speech models (PSMs), step 3 involves using frozen PSMs to extract features, and step 4 involves using score-based assessment methods to determine the best model for the targeted task or the best layer for partial tuning or insertion of parameter-efficient modules.   

\vspace{-2pt}

More specifically, our model assessment framework is based on two  theoretical approaches: (i) optimal transport with latent space measurement~\cite{yang2021voice2series} and (ii) maximum evidence~\cite{you2021logme} with Bayesian probabilistic graphical models (PGMs). Since both of these theoretical approaches have not been extensively explored in speech processing, we make a very first attempt to approximately model continuous speech recognition and isolated word recognition as classification problems by establishing a temporal independent hypothesis~\cite{graves2006connectionist} (TIH). In the following paragraphs, we first introduce recent work on model transferability estimation in non-speech processing tasks and then explain how the TIH connects to the two theoretical backbones, even in a simplified and non-autoregressive decoding setting.

\textbf{Related Pre-Trained Model Transferability Works:}
One of the earliest works on pre-trained model transferability estimation in computer vision tasks was the use of negative conditional entropy (NCE)~\cite{tran2019transferability}.
NCE uses a pre-trained vision model and evaluates
conditional entropy between target pseudo labels, which are the assigned label of the source model and real target
labels. 
A subsequent work to NCE is the log expected empirical predictor (LEEP)~\cite{nguyen2020leep}, which modifies NCE by using soft predictions from the source model. LEEP calculates the log-likelihood between the target labels and the predictions to estimate model transferability, which allows the input data to come from arbitrarily different distributions. However, when evaluating the quality of model transferability estimation, both NCE and LEEP have been reported with correlation coefficient $\leq 0.6$ with high $p$-value, as noted in~\cite{you2021logme}.

On the other hand, a recent model transferability assessment solution for vision and language processing tasks is LogME~\cite{you2021logme}. LogME predicts accuracy on the target task by estimating the marginalized likelihood of labeled target examples, under the premise that a linear classifier is added on top of the pre-trained model. LogME can be applied to both supervised and unsupervised learning tasks for cross-task adaptation, as it considers zero-shot features encoded by pre-trained models.


In recent works~\cite{bassignana2022evidence, agostinelli2022transferability, agostinelli2022stable}, LogME has been shown to be an appropriate first step for encoder selection in \textit{natural language processing}~\cite{bassignana2022evidence}.
Preliminary attempts in~\cite{agostinelli2022transferability} also showed high correlation results when using a modified version of LogME for ensemble selection of large language models. Meanwhile, to the best of the authors' knowledge, there are no studies of using the aforementioned log-likelihood methods to analyze speech processing-based tasks from large-scale pre-trained FM(s) to adaptation. While classification scores have been used as evaluation metrics to assess deep learning models~\cite{nguyen2020leep, you2021logme}, they may not fully capture the performance of speech models due to the unique characteristics of speech signals. The sequential nature of speech signals requires modeling time-step dependencies, which cannot be directly assessed by classification metrics. Next, we review some related speech processing methods and empirical studies to highlight the differences between existing methods and the proposed model assessment framework in speech processing. 





\textbf{Pre-trained Speech Model (PSM) Selection Works:}
In the realm of neural network-based PSM for various speech application tasks, previous research has explored the possibility of selectively reusing established PSMs by only fine-tuning certain trainable components, such as the acoustic encoder or language decoder~\cite{yang2021superb}. For instance, prior research~\cite{zhao2021addressing} suggests that tuning only selected encoders of RNN-Transducer~\cite{graves2012sequence}  closest to the \textit{input layer} of PSM results in improved performance compared to fully fine-tuning. In contrast, \textit{Pasad et al.}~\cite{pasad2021layer} discovered that re-initializing the transformer layers closest to the \textit{prediction head} outperforms initializing all layers from a pre-trained SSL-based PSM. Given that different PSMs feature distinct encoder-decoder architectures that incorporate both acoustic and language information, the selection of an appropriate fine-tuning configuration can be task-dependent and model-specific, requiring heuristic search. In other words, training a model using every possible encoder or decoder has shown an extremely large sample complexity and a significant amount of time~\cite{thomas2022efficient, bassignana2022evidence, zhao2022improving}, making it difficult to reproduce for researchers with limited computational resources.

A preliminary attempt to estimate pre-trained model transferability was proposed in~\cite{yang2021voice2series}. This attempt is based on optimal transport and measures the distance between a target domain and its source domain in terms of population risk, as its difficulty model transferability. However, their work was focused on \textit{selecting} PSMs for \textit{cross-modal} adaptation to classify sensor data, and its effectiveness in speech processing tasks has not been studied. In this work, we also advance the use of optimal transport~\cite{yang2021voice2series,alvarez2020geometric} for estimating PSMs and provide some initial explorations for speech processing, which is based on a proposed temporal independent condition discussed in Section~\ref{2:1}.

The recent advent of FMs and their breakthroughs in speech processing tasks~\cite{ravanelli2020multi,baevski2020wav2vec,ling2020decoar, hsu2021hubert, zhang2022bigssl} has created a growing need for new transferability estimation techniques evaluating the strengths of FMs. 
But speech tasks have yet to be extensively investigated with transferability estimation techniques.
In this work, we aim to answer the question of how well we can estimate the transferability of pre-trained speech models to specific speech recognition tasks. We establish baselines for layer-wise and model-wise evaluations. The proposed framework could serve as one first attempt to evaluate pre-trained speech models, which has some potential to enhance the development of model tuning for future studies.



\textbf{Our contributions include:}
\begin{itemize}
    \item We connect model transferability estimation to speech tasks by leveraging a simplified hypothesis of temporal independence to relaxes the posterior nature of speech recognition.
    \item We advance two different perspectives on model transferability for speech processing: (i) optimal transport and (ii) evidence maximization, to provide interpretable scores for assessing PSMs. 
    \item By conducting initial attempts in cross-layer and cross-model transfer learning setups, we evaluate that our framework has a high rank correlation and low $p$-value.

\end{itemize}

\section{Transferability for Speech Models}
This study proposes two perspectives on evaluating model transferability in speech processing. We explore the use of optimal transport and evidence maximization as methods for generating interpretable scores to assess the performance of speech processing models (PSMs). To enable the evaluation of model transferability, we introduce a simplified hypothesis of temporal independence in Section~\ref{2:1}, which relax the posterior nature of speech processing in this section and enables evaluating model transferability in sequential speech data. In Section~\ref{2:2}, we incorporate TIH into the optimal transport sliced Wasserstein Distance, which is utilized for measuring the distance between source and target data distributions. In Section~\ref{2:3}, we apply TIH in the likelihood aspect, LogME, a technique that models the relationship between extracted features of a speech signal and the output labels.

\subsection{Modeling Temporal Independent Hypothesis (TIH)}
\label{2:1}

To estimate the transferability for speech processing asks, we estimate the correlation between the label sequence and the features extracted from the input sequence. Since in speech processing tasks, the length of the input sequence and the label are not aligned, a temporal independent hypothesis was introduced in the connectionist temporal classification (CTC) mechanism~\cite{graves2006connectionist} for automatic speech recognition (ASR) modeling. It is based on a simplified condition that ignores posterior information during the loss computation. This simplified condition assumes that a speech model can learn sequential information through its neural network encoders, such as attention or recurrent networks.

Let $f_i \in \mathbb{R}^D$ be the $i$-th feature extracted with $D$ dimensions by a pre-trained model $\phi$ for the input $x$, and let $\boldsymbol{y} _i \in \mathbb{Z}^+$ be the scalar label. The collection of all features is represented by a matrix $F \in \mathbb{R}^{D \times n}$, and the collection of all labels is represented by the vector $\boldsymbol{y} \in \mathbb{Z}^n$:

\begin{equation}
    p(\boldsymbol{y}|F)=\sum_{A\in A_F}^{}\prod_{t=1}^{T}p_t(a_t|F),
    \label{eq:ctcalignment}
\end{equation}
where $A$ is defined as the set of all valid alignments between the input $\boldsymbol{x}$ and the output $\boldsymbol{y}$, and $P(\boldsymbol{a} \mid \boldsymbol{x})$ is the probability of a specific alignment $\boldsymbol{a}$ given the input $\boldsymbol{x}$. The probability density $p(\boldsymbol{y}|F)$ measures the compatibility between the feature matrix $F$ and the label sequence $\boldsymbol{y}$. It is calculated by summing over all possible alignments $A_F$ between the input and output sequences, using Equation \ref{eq:ctcalignment}. This equation multiplies the probability of the alignment at each time step $t$, given the feature matrix $F$ and the label sequence $\boldsymbol{y}$.


\textbf{Forced Alignment:} In the case when the input sequence and the output label are not aligned, we estimate the transferability of pre-trained models using the CTC  forced alignment algorithm by aligning the model's output with the reference transcriptions of the target task. Our  forced alignment process includes backtracking
, which enables the correction of misaligned frames by considering previous frames and making necessary adjustments to the current alignment.


\vspace{-10pt}
\begin{figure}[ht!]
    \centering
\includegraphics[width=0.45\textwidth]{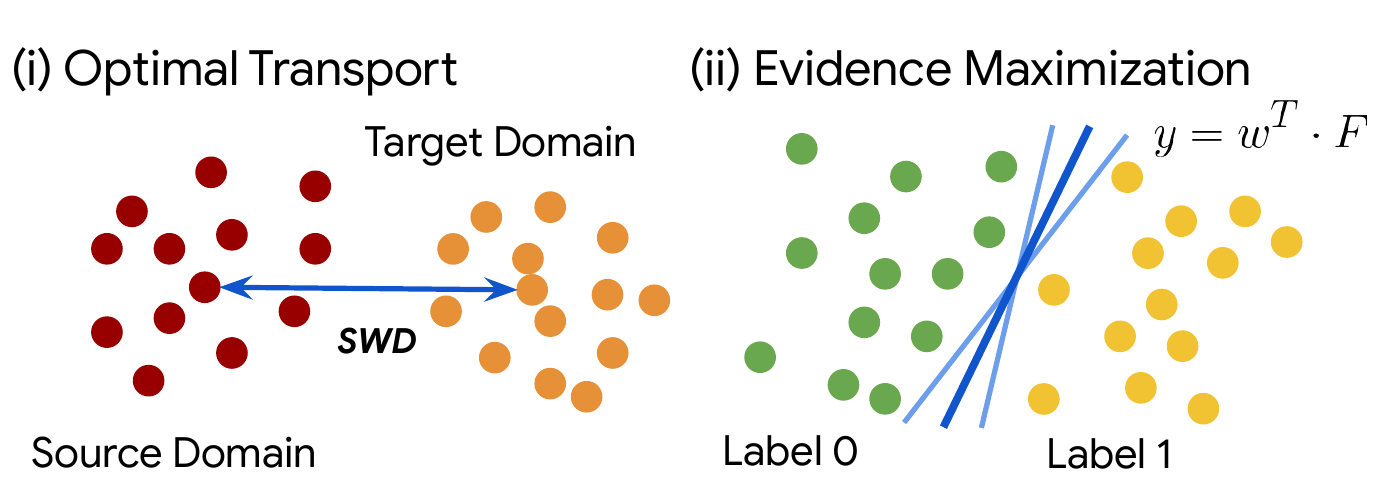}
    \caption{Illustration of the two approaches for estimating transferability in speech processing tasks: optimal transport (Section~\ref{2:2}) and maximum evidence (Section~\ref{2:3}). The transferability metric in~\ref{2:2} is SWD~\cite{yang2021voice2series}, while in~\ref{2:3}, we use LogME ($\log p(\boldsymbol{y}|F) / n$), where $p(\boldsymbol{y}|F)=\int p(w)p(\boldsymbol{y}|F,w)dw$, to assess transferability.}
    \label{fig:bound}
\end{figure}
\vspace{-15pt}
\subsection{Optimal Transport by Sliced Wasserstein Distance}
\label{2:2}

In speech processing tasks, the output label and input sequence may not align, directly calculating the median distance of the data distribution is infeasible. So here, we introduce optimal transport with TIH.
Building upon Eq.~(\ref{eq:ctcalignment}), we conduct latent space measurement by sampling a random batch of source ($\boldsymbol{x}_t^{\text{src}}$) and target inputs ($\boldsymbol{x}_t^{\text{trg}}$) for time step $t$. We model character-level prediction using the TIH and the latent distribution of $\boldsymbol{\mu}_t^{\text{src}}$ and $\boldsymbol{\mu}_t^{\text{trg}}$, which represent the zero-shot latent representation. We measure the distance between these two probability distributions, $\boldsymbol{\mu}_t^{\text{src}}$ and $\boldsymbol{\mu}_t^{\text{trg}}$, using the sliced Wasserstein distance~\cite{kolouri2019generalized} (SWD), $\mathcal{W}_p(\boldsymbol{\mu}_t^{\text{src}},\boldsymbol{\mu}_t^{\text{trg}})$, under  $L_p$-norm, where $p=1$ is used to obtain a closed form solution based on the theory presented in~\cite{yang2021voice2series}.
A larger SWD value (or score) indicates a greater difficulty in aligning the latent representation between the two selected domains and serves as a measure of the difficulty of model adaptation. In practice, we calculate the median SWD value over the total number of time steps, $T$, and report this value as its transferability score. However, this TIH-based SWD score \textbf{cannot} be used directly for evaluating SSL tasks as the unsupervised pre-trained data and target tasks come from the same domain (e.g., Librispeech~\cite{panayotov2015librispeech}). In the next section, we investigate a more general framework for estimating the transferability of PSMs.


\subsection{Transferability Estimation by Likelihood}
\label{2:3}
We illustrate a connection between the aforementioned TIH and likelihood-based estimation in LogME~\cite{you2021logme} by computing the probability of the target labels conditioned on the source features. LogME measures the suitability of the encoded features for predicting labels via a probability density, which is estimated by mapping $F$ to $\boldsymbol{y}$ using a linear transformation parameterized by $w$. The goal is to find the optimal weight  $w^*$: 
\begin{equation}
   p(\boldsymbol{y}|F) \rightarrow
 p(\boldsymbol{y}|F,w^*)\rightarrow\int p(w)p(\boldsymbol{y}|F,w)dw,
\label{eq:logme}
\end{equation}
where Eq. (\ref{eq:logme}) denotes a proxy for feature suitability.


Next, the probability densities of $w$ and $p(\boldsymbol{y}|F, w)$ are modeled using positive hyper-parameters $\alpha \in \mathbb{R}^+$ and $\beta\in \mathbb{R}^+$. The prior distribution of the weights is modeled as an isotropic multivariate Gaussian $w \sim \mathcal{N}(0, \alpha^{-1} I)$, and each observation's distribution is modeled as a one-dimensional normal distribution $p(\boldsymbol{y}_i | f_i, w, \beta) \sim \mathcal{N}(\boldsymbol{y}_i | w^Tf_i, \beta^{-1})$, which is computed by: 
\begin{equation}
    p(\boldsymbol{y}|F, \alpha, \beta) = \int p(w|\alpha)p(\boldsymbol{y}|F, w, \beta) dw
\label{eq:logme-hyperparam}
\end{equation}
The logarithm maximum evidence  $\mathcal{L}(\alpha^*, \beta^*) = \log p(\boldsymbol{y}|F, \alpha^*, \beta^*)$
is used to evaluate the compatibility between features and labels. $\mathcal{L}(\alpha^*, \beta^*) / n$ is used as our LogME evaluation metric. 


Based on Equations (\ref{eq:ctcalignment}) and (\ref{eq:logme-hyperparam}), we propose a new formulation of LogME for speech processing tasks. The model is parameterized by hyper-parameters $\alpha$ and $\beta$ and is given by the following equation:
\begin{equation}
p(\boldsymbol{y}|F, \alpha, \beta) \rightarrow \sum_{A\in A_F}^{}\prod_{t=1}^{T}\int p_t(w_t|\alpha)p(a_t|F,w_t, \beta)dw_t.
\end{equation}

We compute the log-likelihood of the labels in the aligned sequence and sum them to obtain the LogME score for the speech model. It is important to note that this revised LogME measurement can be used to estimate both supervised speech model training with labeled data and SSL models, without a need for  source data.


\section{Experiments}
In our speech task experiments, we conduct layer-wise exploration to estimate the transferability of pre-trained models. 

\subsection{Experimental setup}
\textbf{Continuous ASR}: Our experimental setup follows the same training procedure as reported in~\cite{yang2023english} using the official test data. We conduct ASR adaptation experiments using a pre-trained English-only Conformer model~\cite{gulati2020conformer}.
We evaluate the average word error rate (WER) on the Multilingual Librispeech (MLS) benchmark, which includes seven languages~\cite{yang2023english, li2023efficient}. The supervised Conformer-ASR model was trained with a batch size of 1024 on 64 TPUs following the same setup as in~\cite{yang2023english}. 
\textbf{Phoneme and Isolated Word Recognition}: We evaluated the performance of our models on the LibriSpeech dataset using the train-clean-100, dev-clean, and test-clean subsets. The evaluation metric was Phone Error Rate (PER) in Table~\ref{tab:2:per}. For isolated word recognition tasks, we used the Speech Commands dataset v1.0~\cite{warden2018speech}, which includes ten keyword classes, silence, and an unknown class. The evaluation metric for this task was accuracy (ACC) in Tables \ref{tab:3} and~\ref{tab:4}. In the SSL setup, the raw waveform was used as input, and a linear head was added to the upstream SSL model for phoneme recognition and speech command tasks. The SSL models were trained on a single RTX3090 GPU using the experimental settings from the SUPERB benchmark~\cite{yang2021superb}.
\textbf{A tSNE-based Baseline Method:}
The tSNE-based clustering~\cite{van2008visualizing} has been commonly used for model interpretability.
We also calculate the median tSNE clustering points of the unpaired source ($\boldsymbol{x}_t^{\text{src}}$) and target input ($\boldsymbol{x}_t^{\text{trg}}$) distributions. As an extra baseline for transferability estimation, this tSNE score is similar to the SWD score, where a large distance indicates greater difficulty in aligning the two different representations.


\subsection{Study 1: Layer-Wise Exploration}

\vspace{-4pt}
\begin{table}[ht!]
\caption{Fine-Tuning pre-trained English-only Conformer for Multilingual Librispeech (MLS)~\cite{pratap2020mls} for cross-lingual adaptation, which WER is reported as averaged of seven MLS languages as the setup in~\cite{yang2023english}.}
\label{tab:conf:mls}
\centering
\begin{adjustbox}{width=0.50\textwidth}
\begin{tabular}{ccc|c|c|c}
\toprule
\multicolumn{1}{c|}{RNN-T Layer} &
  \multicolumn{1}{c|}{WER ($\downarrow$)} &
   Rank$_\text{FT}$ &
  Rank$_\text{tSNE}$ &
 Rank$_\text{LogME}$ &
  \multicolumn{1}{c}{Rank$_\text{SWD}$} \\ \hline
\multicolumn{1}{c|}{Conf-01} &  \multicolumn{1}{c|}{62.63} & 17 & 17    & 17       & 17      \\ 
\multicolumn{1}{c|}{Conf-02} & \multicolumn{1}{c|}{53.49} & 15 & 16    & 10       & 11      \\ 
\multicolumn{1}{c|}{Conf-03} & \multicolumn{1}{c|}{53.61} & 16 & 15    & 15       & 10      \\ 
\multicolumn{1}{c|}{Conf-04} & \multicolumn{1}{c|}{47.75} & 14 & 13    & 13       & 14      \\ 
\multicolumn{1}{c|}{Conf-05} & \multicolumn{1}{c|}{37.02} & 11 & 14    & 16       & 13      \\ 
\multicolumn{1}{c|}{Conf-06} & \multicolumn{1}{c|}{48.71} & 13 & 12    & 12       & 16      \\ 
\multicolumn{1}{c|}{Conf-07} & \multicolumn{1}{c|}{42.13} & 12 & 5     & 14       & 15      \\ 
\multicolumn{1}{c|}{Conf-08} & \multicolumn{1}{c|}{32.32} & 10 & 3     & 6        & 8       \\ 
\multicolumn{1}{c|}{Conf-09} & \multicolumn{1}{c|}{21.74} & 7  & 1     & 7        & 9       \\ 
\multicolumn{1}{c|}{Conf-10} & \multicolumn{1}{c|}{22.56} & 8  & 10    & 9        & 4       \\ 
\multicolumn{1}{c|}{Conf-11} & \multicolumn{1}{c|}{19.86} & 4  & 4     & 5        & 6       \\ 
\multicolumn{1}{c|}{Conf-12} & \multicolumn{1}{c|}{21.71} & 6  & 8     & 11       & 12      \\ 
\multicolumn{1}{c|}{Conf-13} & \multicolumn{1}{c|}{25.56} & 9  & 7     & 8        & 7       \\ 
\multicolumn{1}{c|}{Conf-14} & \multicolumn{1}{c|}{19.23} & 3  & 9     & 4        & 5       \\ 
\multicolumn{1}{c|}{Conf-15} & \multicolumn{1}{c|}{20.09} & 5  & 11    & 3        & 2       \\ 
\multicolumn{1}{c|}{Conf-16} & \multicolumn{1}{c|}{18.87} & 2  & 6     & 2        & 3       \\ 
\multicolumn{1}{c|}{Conf-17} & \multicolumn{1}{c|}{18.27} & 1  & 2     & 1        & 1       \\ \hline
\multicolumn{3}{c|}{Spearman's rank correlation coefficient ($\uparrow$)}                                   & $0.69$  & $0.87$     & $0.81$    \\ 
\multicolumn{3}{c|}{$p$-value ($\downarrow$)}                                                                   & $1\times10^{-3}$ & $6\times10^{-6}$ & $7\times10^{-5}$ \\ \bottomrule
\end{tabular}
\end{adjustbox}
\end{table}
\vspace{-3pt}
\begin{table}[ht!]
\caption{HuBERT-based SSL PSM layer-wise fine-tuning results}
\centering
\begin{adjustbox}{height = 67pt,width=0.35\textwidth}
\begin{tabular}{@{}c|c|c|c@{}}

\toprule
HuBERT Layer                       & PER ($\downarrow$)                  & Rank$_\text{FT}$                 & Rank$_\text{LogME}$      \\\hline
Layer-01                   & 35.63                        & 12                      & 9                            \\
Layer-02                   & 29.61                        & 9                       & 10                           \\
Layer-03                   & 27.51                        & 8                       & 8                            \\
Layer-04                   & 25.43                        & 7                       & 7                            \\
Layer-05                   & 23.75                        & 6                       & 6                            \\
Layer-06                   & 18.83                        & 5                       & 5                            \\
Layer-07                   & 14.35                        & 4                       & 4                            \\
Layer-08                   & 10.86                        & 3                       & 2                            \\
Layer-09                   & 8.73                        & 2                       & 3                            \\
Layer-10                   & 7.40                         & 1                       & 1                            \\
Layer-11                  & 29.84                        & 10                      & 12                           \\
Layer-12                  & 30.37                        & 11                      & 11                           \\\hline
\multicolumn{3}{c|}{ Spearman's rank correlation coefficient~($\uparrow$)} & \multicolumn{1}{c}{$0.94$}    \\
\multicolumn{3}{c|}{$p$-value~($\downarrow$)}                                                        & \multicolumn{1}{r}{$3\times10^{-6}$}
\\\bottomrule
\end{tabular}
\end{adjustbox}
\label{tab:2:per}
\end{table}

Transferability estimation can aid in determining the optimal layer for fine-tuning pre-trained models. In Study 1, we performed a layer-wise analysis to investigate the fine-tuning of a pre-trained English-only Conformer for MLS cross-lingual adaptation~\cite{pratap2020mls}, as detailed in Table \ref{tab:conf:mls}. Our results suggest that fine-tuning the top layers achieves better performance. Using LogME, we evaluated the performance of fine-tuning each layer, obtaining a correlation of 0.87 with a $p$-value of $6\times10^{-6}$. Additionally, SWD accurately estimated the ranking with a correlation of 0.81 and a $p$-value of $7\times10^{-5}$. However, direct computation of distance with tSNE was found to be less precise. Our experimental results demonstrate that layer-wise analysis using the LogME score is a more accurate method for estimating transferability in speech processing tasks, which provide insights into selecting the optimal layer for fine-tuning in cross-lingual ASR transfer.


In SSL models, tSNE and SWD cannot be used to estimate transferability as they rely on source data, which is not available in this task. To address this issue, we performed a layer-wise analysis of a HuBERT base model~\cite{hsu2021hubert} for phoneme recognition. Our results, presented in Table~\ref{tab:2:per}, indicate that the LogME score is a precise method for evaluating transferability in speech processing tasks. Our findings also suggest that fine-tuning the top layers does not necessarily result in improved performance, which differs from the results of the RNN-T model presented in Table~\ref{tab:conf:mls}. Therefore, estimating transferability for individual layers is essential, as it can save time and computational costs and eliminate the need for expert knowledge in selecting the optimal layers for fine-tuning.


\subsection{Study 2: Model-Wise Exploration}
\subsubsection{Pre-trained speech model tuned on classification}

For model-wise exploration, we conducted experiments to explore the transferability of pretrained models tuned on the classification tasks. To this end, we evaluated six different pre-trained models and computed their transferability scores using various evaluation metrics. As presented in Table~\ref{tab:3}, all of the evaluation metrics produced accurate estimates of the transferability scores. Our experimental results thus demonstrate that the LogME score can effectively capture the transferability of a pre-trained model in speech classification tasks.


\subsubsection{Different pre-train data on same neural architecture}
We evaluated transferability metrics on the AST model \cite{gong2021ast} using three pre-trained datasets: frozen +Linear Probing, ImageNet \cite{deng2009imagenet}, and AudioSet \cite{gemmeke2017audio}, to examine the effectiveness of transferability metrics evaluation on different pre-training data. Only the last MLP layer was fine-tuned during the experiments. The results in Table~\ref{tab:4} indicate that the LogME score effectively estimates the transferability of pre-trained models in speech classification tasks. The model pre-trained on AudioSet has the best transferability, whereas the model with random initialization showed the worst transferability. These findings underscore the significance of selecting appropriate pre-training data to enhance model transferability.

\begin{table}[ht!]
\centering
\caption{Evaluation on GSC-v1~\cite{warden2018speech} on different PSMs.}
\begin{adjustbox}{width=0.5\textwidth}
\begin{tabular}{lrccccc}
\toprule
Models & Para. & \multicolumn{1}{l}{Acc.~($\uparrow$)} & \multicolumn{1}{c}{Rank$_\text{FT}$}     & \multicolumn{1}{c}{Rank$_\text{LogME}$}  & \multicolumn{1}{c}{Rank$_\text{tSNE}$} & \multicolumn{1}{c}{Rank$_\text{SWD}$}    \\ \hline
$^\dag$HuBERT \cite{hsu2021hubert}   &  95M  & 95.94 & 2 & 2 & 1 & 2 \\
$^\dag$Wav2Vec2 \cite{baevski2020wav2vec}&  95M    & 92.27 & 5 & 5 & 5 & 5 \\
$^\dag$DeCoAR2.0 \cite{ling2020decoar}  &  90M    & 92.63 & 4 & 4 & 4 & 4 \\
$^\dag$Vggish~\cite{hershey2017cnn}   & 72M  & 96.78 & 1 & 1 & 2 & 1 \\
Yamnet~\cite{fonseca2019audio}   & 4M & 94.32 & 3 & 3 & 3 & 3 \\
fsFCNN~\cite{hu2020device}  & 20M  & 91.34 & 6 & 6 & 6 & 6 \\ \hline
\multicolumn{3}{c}{Time$^\dag$}                    & \multicolumn{1}{r}{$\sim0.61$Day} & \multicolumn{1}{r}{$24.09$s} & \multicolumn{1}{r}{$28.01$s}        & \multicolumn{1}{r}{$10.86$s} \\ \hline
\end{tabular}
\end{adjustbox}
\label{tab:3}
\end{table}
\vspace{-4pt}
\begin{table}[ht!]
\centering
\caption{Same AST~\cite{gong2021ast} as PSM but different pre-trained setups. }
\begin{adjustbox}{width=0.40\textwidth}
\begin{tabular}{@{}ccc@{}}
\toprule
PSM Setup & FT Acc. ($\uparrow$)  & LogME$_\text{TIH}$\\ \midrule
ImageNet + Audioset~\cite{gemmeke2017audio}    & 92.37       & 0.6483      \\
ImageNet~\cite{deng2009imagenet}    & 89.72       & 0.5668      \\
Frozen + Linear-probing      & 32.56       & -0.1247     \\ \hline
Time            & $214$s      & $1.75$s        \\
\bottomrule
\end{tabular}
\end{adjustbox}
\label{tab:4}
\end{table}

\vspace{-3pt}
\section{Conclusion}
In this study, we explored the transferability of pre-trained upstream models in speech recognition tasks. Our results showed we are able to estimate the transferability of speech processing tasks. Moreover, our model-wise analysis showed that different pre-trained upstream models had varying adaptability to different downstream tasks. We believe our findings can provide insights into improving the efficiency and effectiveness of transfer learning in speech recognition. Code is available at: \url{github.com/virginiakm1988/LogME-CTC}.

\label{section:preprints}
\clearpage
\bibliographystyle{IEEEtran}
\bibliography{mybib}
\clearpage



\end{document}